\documentclass[final]{cvpr}

\usepackage{times}
\usepackage{epsfig}
\usepackage{graphicx}
\usepackage{amsmath}
\usepackage{amssymb}
\usepackage{verbatim}
\usepackage{bm}
\usepackage[pagebackref=true,breaklinks=true,letterpaper=true,colorlinks,bookmarks=false]{hyperref}

\usepackage{multirow}
\usepackage{booktabs}
\usepackage{subfig}

\newcommand{\tablestyle}[2]{\setlength{\tabcolsep}{#1}\renewcommand{\arraystretch}{#2}\centering\small}

\newlength\savewidth\newcommand\shline{\noalign{\global\savewidth\arrayrulewidth
  \global\arrayrulewidth 1pt}\hline\noalign{\global\arrayrulewidth\savewidth}}

\usepackage[pagebackref=true,breaklinks=true,colorlinks,bookmarks=false]{hyperref}

\begin{document}

\title{Self-Supervised Learning for Semi-Supervised Temporal Action Proposal}

\author{
    Xiang Wang$^1$
    \hspace{0.2cm} Shiwei Zhang$^2$
    \hspace{0.2cm} Zhiwu Qing$^1$
    \hspace{0.2cm} Yuanjie Shao$^1$\thanks{Corresponding Author}
    \hspace{0.2cm} Changxin Gao$^1$ 
    \hspace{0.2cm} Nong Sang$^1$\\[.5ex]
    $^1$Key Laboratory of Image Processing and Intelligent Control,\\ School of Artificial Intelligence and Automation, \\Huazhong University of Science and Technology, China\\
    \hspace{0.3cm} $^2$DAMO Academy, Alibaba Group, China
    \\
    {\tt\small \{wxiang,qzw,shaoyuanjie,cgao,nsang\}@hust.edu.cn, zhangjin.zsw@alibaba-inc.com}
}

\maketitle
\pagestyle{empty}  
\thispagestyle{empty} 

\begin{abstract}

Self-supervised learning presents a remarkable performance to utilize unlabeled data for various video tasks.
In this paper, we focus on applying the power of self-supervised methods to improve semi-supervised action proposal generation.
Particularly, we design an effective \textbf{S}elf-supervised \textbf{S}emi-supervised \textbf{T}emporal \textbf{A}ction \textbf{P}roposal (SSTAP) framework.
The SSTAP contains two crucial branches, i.e., temporal-aware semi-supervised branch and relation-aware self-supervised branch.
The semi-supervised branch improves the proposal model by introducing two temporal perturbations, i.e., temporal feature shift and temporal feature flip, in the mean teacher framework.
%
The self-supervised branch defines two pretext tasks, including masked feature reconstruction and clip-order prediction, to learn the relation of temporal clues.
By this means, SSTAP can better explore unlabeled videos, and improve the discriminative abilities of learned action features.
%
%
We extensively evaluate the proposed SSTAP on THUMOS14 and ActivityNet v1.3 datasets. The experimental results demonstrate that SSTAP significantly outperforms state-of-the-art semi-supervised methods and even matches fully-supervised methods. 
%
Code is available at \url{https://github.com/wangxiang1230/SSTAP}.

\end{abstract}

\section{Introduction}

Temporal action proposal aims to localize action instances in untrimmed videos by predicting both action-ness probabilities and temporal boundaries.
%
%
Recently, various approaches~\cite{CTAP,BMN,DBG} for the task have been proposed and achieve significant progress with the quick development of spatio-temporal feature learning~\cite{2014Two, C3D, I3D, slowfast}.
Almost all the methods rely on dense temporal annotations for the training videos.
However, the annotating task is tedious and requires large amounts of human labor.
Thus these methods may have limited abilities to meet practical demands.
\begin{figure}[t]
\begin{center}
\includegraphics[width=0.99\linewidth]{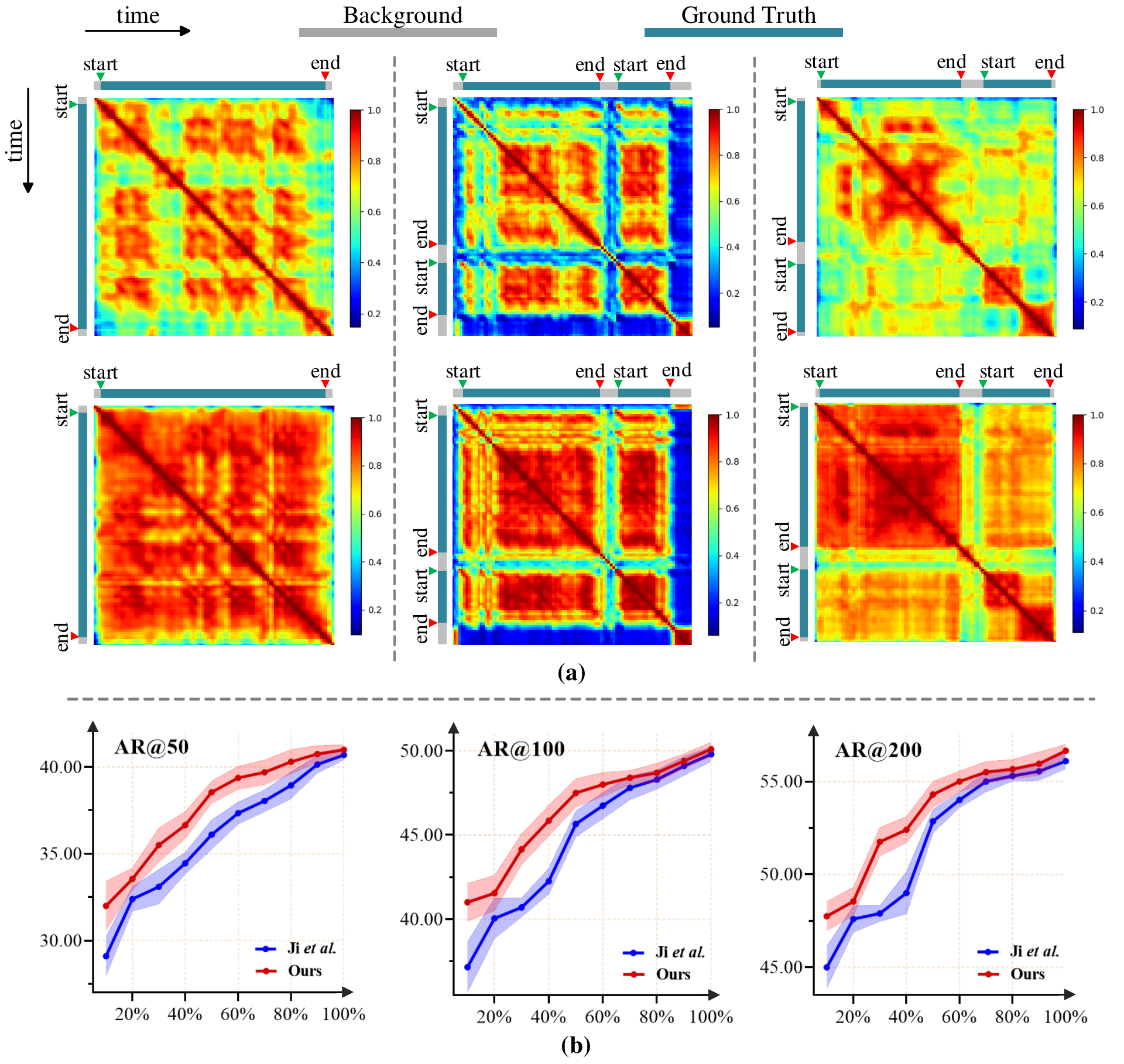}
\vspace*{-4mm}
\end{center}
\vspace{-6pt} 
   \caption{\textbf{(a)} Feature similarity matrix visualization. We use cosine similarity to measure the degree of similarity between arbitrary two snippet-level feature vectors within the same video. Note that, snippet-level features of the action as similar as possible while separating actions from backgrounds. Compared to Ji \etal~\cite{ji-Semi} (top), better representations of the features can be learned by adding our relation-aware self-supervised branch (bottom). \textbf{(b)} 
   Our SSTAP consistently exceeds the state-of-the-art semi-supervised method (Ji \etal~\cite{ji-Semi}) in terms of Average Recall when trained with different percentages of labels on the THUMOS14 dataset.
   }
\label{fig:introduce}
\vspace*{-5mm}
\end{figure}
%
%

To alleviate the dependence of labeled videos, Ji~\etal~\cite{ji-Semi} first apply the semi-supervised method, \emph{i.e.}, Mean Teacher~\cite{mean-teacher}, to temporal action proposal.
In this method, Ji \etal only use a small portion of labeled videos and reach high performances.
Due to perturbation is an essential component of semi-supervised methods, the method proposes two sequential perturbations, \emph{i.e.}, time warping and time masking, to improve robustness and generalization.
However, the perturbations ignore the temporal interactions, which is critical to learn robust action representations. 
Another line of paradigm to utilize unlabeled videos is about self-supervised methods.
These methods explore undergoing video structure by predefining pretext tasks, \emph{e.g.}, learning temporal order~\cite{self-sort-order, self-clip-order}, 
pace prediction~\cite{self-pace}, and learning playback rate~\cite{self-PRP}.
They have reached an impressive performance in several video-related tasks~\cite{self-tschannen2020self, self-sort-order,self-temporal}, and thus self-supervised learning is proved to be a promising methodology.
However, the methodology has never been explored to generate temporal action proposals.
We believe that it can contribute to improving the performance by fully utilizing unlabeled videos.

Based on the above observations, we propose to apply self-supervised methods to improve the semi-supervised temporal action proposal by designing the SSTAP framework.
The proposed SSTAP contains two main branches, \emph{i.e.}, temporal-aware semi-supervised branch and relation-aware self-supervised branch.
The temporal-aware semi-supervised branch targets to improve the method in~\cite{ji-Semi} by designing two simple but effective perturbations, \emph{i.e.}, temporal feature shift and temporal feature flip.
The first perturbation bidirectionally moves some randomly selected channels of feature maps, which is inspired by~\cite{TSM}, and the second perturbation flips the total features, both of them along the temporal dimension.
By this means, the proposal model can be more robust and generalized.
In the relation-aware self-supervised branch, we define two pretext tasks, \emph{i.e.}, masked feature reconstruction and clip-order prediction.
%
%
The pretext tasks respectively reconstruct the randomly masked features and predict the correct order of the randomly shuffled clip features.
Therefore, SSTAP can better explore the unlabeled videos and learns discriminative features.
In Figure~\ref{fig:introduce}(a), the covariance-like similarity matrixes show that the self-supervised branch can help to decrease intra-class distance and increase inter-class distance simultaneously. Hence SSTAP can improve proposal performance (Figure~\ref{fig:introduce}(b)).
We evaluate the proposed SSTAP on the challenging THUMOS14~\cite{thumos14} and ActivityNet v1.3~\cite{activitynet} datasets and achieve a remarkable improvement on both datasets.

In summary, our main contributions are as follows:
\begin{itemize}
\item[$\bullet$] To the best of our knowledge, we are the first to incorporate self-supervised learning in semi-supervised temporal action proposal by designing a unified SSTAP framework;
\item[$\bullet$] We have designed two simple but effective types of temporal sequential perturbations and defined two types of self-supervised pretext tasks for SSTAP;
\item[$\bullet$] We extensively test the proposed SSTAP on two public datasets and achieve state-of-the-art performance.
%
\end{itemize}

\begin{figure*}
\begin{center}
\includegraphics[width=0.80\linewidth]{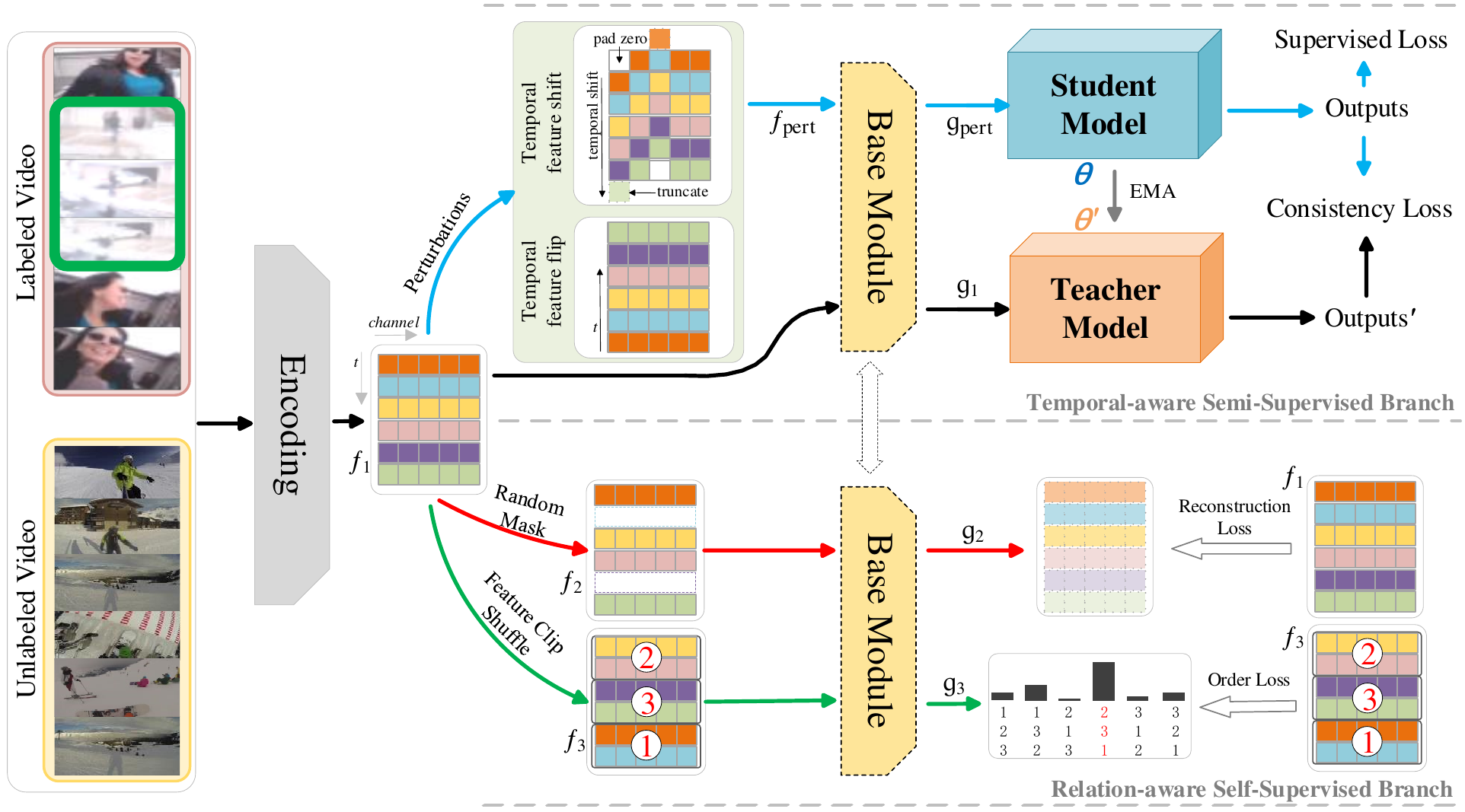}
\vspace*{-6mm}
\end{center}
   \caption{Overview of our SSTAP. We first encode a sampled untrimmed input video into a feature sequence $f_{1}$. In the temporal-aware semi-supervised branch (top right), there are two sequential perturbation operations: temporal feature shift and temporal feature flip. And the Base Module takes the perturbed sequences $f_{pert}$ and the unobstructed $f_{1}$ as inputs. Next, the student model and the teacher model of the same network structure generate outputs. In the relation-aware self-supervised branch (bottom right), there are two self-supervised pretext tasks: masked feature reconstruction and clip-order prediction. In the end, a unified multi-task framework is exploited for optimization. Color-coded arrows denote the associations between the features in the framework and the respective modules.} 
\label{fig:Network}
\vspace*{-4mm}
\end{figure*}
\vspace{-1.5mm}
\section{Related Work}
%
\noindent \textbf{Fully-Supervised Temporal Action Proposal. } 
   There are two mainstream approaches: anchor-based methods and boundary-based methods.
   Anchor-based methods generate proposals by designing multi-scale anchors or sliding windows.
   The works in~\cite{SCNN,DAPs} adopt the C3D network~\cite{C3D} as the binary classifier for sliding window proposal evaluation. 
   The works in~\cite{caba2016fast,SST,SSTAD} use LSTM networks to evaluate the pre-defined anchors.~\cite{TURN,CBR,RC3D,MLTPN,Rethinking} propose to apply temporal regression to adjust the action boundaries.~\cite{CTAP} proposes to use the complementarity of multi-scale anchors and sliding windows to improve performance.
%
%
   Instead, boundary-based methods evaluate each temporal location in the video. TAG~\cite{TAG} generates proposals by a temporal watershed algorithm to group continuous high-score regions. BSN~\cite{BSN} generates proposals via locally locating temporal boundaries and globally evaluating confidence scores. 
   MGG~\cite{MGG} combines anchor-based methods and boundary-based methods to generate proposals. 
   The works in~\cite{GTAD,BCGNN} propose to use graph convolutional networks~\cite{GCN} to model temporal relationships in the input video. BMN~\cite{BMN} proposes a boundary-matching mechanism for the confidence evaluation of densely distributed proposals in an end-to-end pipeline. BMN has become the champion method on ActivityNet Challenge 2019~\cite{activitynet} and the mainstream solution on ActivityNet Challenge 2020~\cite{activitynet}. In this work, we focus on evaluating our SSTAP with the BMN due to its superior performance. 

   \noindent \textbf{Semi-Supervised Learning. } Semi-supervised learning describes a class of algorithms that seek to learn from both unlabeled and labeled data, typically assumed to be sampled from the same or similar distributions. Approaches differ on what information to gain from the structure of the unlabeled data. In the image classification task, there are two important approaches for semi-supervised learning: pseudo-labeling and consistency regularization. Pseudo-label~\cite{pseudo-label} imputes approximate classes on unlabeled data by making predictions from a model trained only on labeled data. Consistency regularization methods measure the discrepancy between predictions made perturbed data points. Approaches of this kind include $\Pi$-Model~\cite{Temporal-ensembling}, Temporal ensembling~\cite{Temporal-ensembling}, Mean Teacher~\cite{mean-teacher}, and Virtual Adversarial Training~\cite{VAT}. In the semi-supervised temporal action proposal task,~\cite{ji-Semi} adopts the Mean Teacher framework and proposes two perturbations.

   \noindent \textbf{Self-Supervised Learning. } Self-supervised learning is a general learning framework that relies on surrogate tasks that can be formulated using only unlabeled data. For image data, there exist self-supervised tasks such as predicting relative positions of image patches~\cite{image-unsupervised-1}, jigsaw puzzles~\cite{image-unsupervised-2}, image inpainting~\cite{image-unsupervised-3} and image color channel prediction~\cite{image-unsupervised-4}. Since the particular property of the video is temporal information, recent works also attempt to leverage the temporal relations among frames, such as order verification~\cite{video-unsupervised-1,video-unsupervised-2}, order prediction of frames~\cite{self-sort-order,self-clip-order}, and perceive multiple temporal resolutions~\cite{self-PRP}.
%
%
%
\section{SSTAP}

   Following the previous work~\cite{ji-Semi}, we build our SSTAP on top of a state-of-the-art fully-supervised proposal generation network, Boundary-Matching Network (BMN)~\cite{BMN}. Note that, compared with the multi-stage BSN~\cite{BSN} framework employed by ~\cite{ji-Semi}, the BMN with end-to-end training can eliminate the mutual influence between multiple stages. At the same time, we also have conducted a fair comparison with~\cite{ji-Semi} using the same BMN as our SSTAP. We extend the Mean Teacher~\cite{mean-teacher} framework with two types of sequential perturbations, \textit{i.e.,} temporal feature shift and temporal feature flip in the temporal-aware semi-supervised branch. And in the relation-aware self-supervised branch, two types of self-supervised auxiliary tasks, \textit{i.e.,} masked feature reconstruction and clip-order prediction, are utilized to assist in training the proposal model. Figure~\ref{fig:Network} shows an overview of our method.
   %
%
\begin{figure}[t]
\begin{center}
\includegraphics[width=0.9\linewidth]{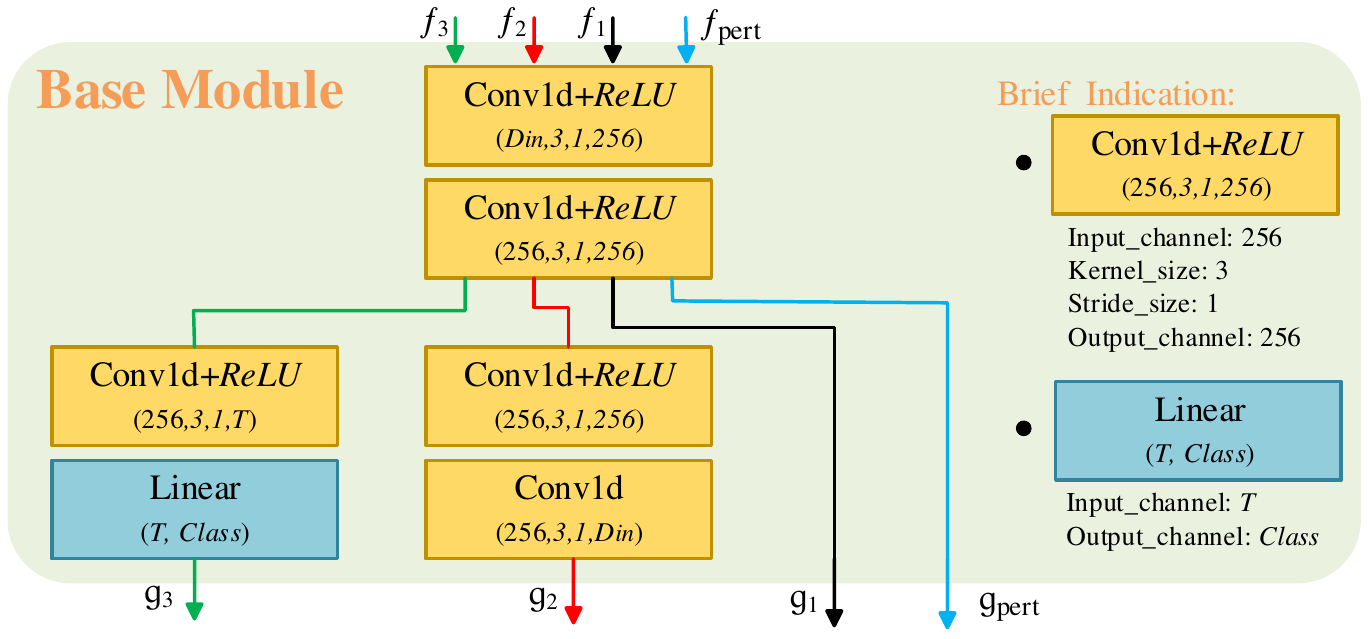}
\vspace{-6mm}
\end{center}
   \caption{The details of our Base Module. Our Base Module is an extension of the ``Base Module'' in BMN~\cite{BMN}}
\label{fig:Base-Module}
\vspace{-4mm}
\end{figure}
\subsection{Problem Description}
   Given an untrimmed video sequence $ S ={\left\{ s_{n} \right \}_{n=1}^{l_{s}}} $ with its length as $l_{s}$, our method aims at detecting action instances $\varphi_{p} = \left\{ \xi_{n} = \left [ t_{s,n}, t_{e,n}\right ] \right \}_{n=1}^{M_{s}} $ with a relatively small amount of training labels, where $M_{s}$ is the total number of action instances, and $\left [ t_{s,n}, t_{e,n}\right ]$ denotes the starting and ending points of an action instance $\xi_{n}$, respectively. Note that, classes of these action instances are not considered in the semi-supervised temporal action proposal task. 
\subsection{Feature Encoding}
   Following recent proposal generation methods~\cite{BSN,BMN,ji-Semi,GTAD}, we construct SSTAP framework upon visual feature sequence extracted from the raw video. Given an untrimmed video sequence $ S ={\left\{ s_{n} \right \}_{n=1}^{l_{s}}} $ with length $l_{s}$, we first divide it into non-overlapping short snippets that contain $\sigma$ frames each. Then the two-stream network~\cite{TSN} is adopted to extract a visual feature sequence $ \mathbf{\phi}={\left\{ \phi_{t_{n}} \right \}_{n=1}^{T}}\in  \mathbb{R}^{ T\times C} $, where $C$ is the dimension of feature and $T = l_{s}/\sigma $.
%
\subsection{Temporal-aware Semi-Supervised Branch}
\label{section:3.2}
   In this section, we present our temporal-aware semi-supervised branch in SSTAP. We first provide a brief description of the proposal generation network and mean teacher framework. Afterward, we introduce two types of sequential perturbations proposed by us, \textit{i.e.,} temporal feature shift and temporal feature flip.
   
\noindent \textbf{Proposal Generation Network.} To validate our semi-supervised framework and better illustrate our approach, we build our method on top of the Boundary-Matching Network (BMN)~\cite{BMN} , an effective and end-to-end proposal generation method. 
   
   The same feature encoding is performed as the first step. The BMN comprises three modules: ``Base Module'', ``Temporal Evaluation Module'' (TEM), ``Proposal Evaluation Module'' (PEM). ``Base Module'' handles the input feature sequence $ \mathbf{\phi}$ and outputs feature sequence $ \mathbf{{\phi}'}$ shared by the following TEM and PEM. TEM evaluates the starting and ending probabilities of each location in the video to generate boundary probability sequences. PEM contains a Boundary-Matching layer (like the ROI Pooling in Faster-RCNN~\cite{faster-rcnn}) to transfer the feature sequence $ \mathbf{{\phi}'}$ to a boundary-matching feature map and contains a series of 3D and 2D convolutional layers to generate boundary-matching confidence maps. The three modules are trained in a unified framework. Therefore, given an untrimmed video, BMN can simultaneously generate (1) boundary probability sequences to construct proposals and (2) boundary-matching confidence maps to evaluate the confidences of all proposals densely. Please refer to~\cite{BMN} for more details of BMN.

%
%
%

\noindent \textbf{Mean Teacher Framework. }
   In the Mean Teacher framework, there are two models: a student proposal model $f_{\theta}$ and a teacher proposal model $f_{{\theta}'}$. The student proposal model learns as in fully-supervised learning, with its weights $\theta$ optimized by the supervised losses applied on labeled videos. The teacher proposal model has the identical neural network architecture as the student, while its weights ${{\theta}'}$ are updated with an exponential moving average (EMA) of the weights from a sequence of student models of different training iterations:
   \begin{equation} \label{eq2}
   {{\theta}'}_{\tau} = \alpha{{\theta}'}_{\tau-1} + \left ( 1-\alpha\right ){\theta}_{\tau},
   \end{equation}
   where $\tau$ denotes the training iteration, and $\alpha$ is a smoothing coefficient, which is always set to 0.999.

\noindent \textbf{Sequential Perturbations. } In the temporal-aware semi-supervised branch, the Mean Teacher framework is adopted on BMN to form our semi-supervised learning framework. Meanwhile, in the literature, stochastic perturbations have been found crucial for learning robust models by many semi-supervised learning works~\cite{Temporal-ensembling,VAT,mean-teacher,Mixmatch,ji-Semi,Enaet}. And a typical way of perturbation is adding noise to feature maps. The work in~\cite{mean-teacher} adds gaussian noise to intermediate feature maps of both student and teacher models. Ji \etal~\cite{ji-Semi} add two perturbations to the input sequence. However, those perturbations ignore the temporal interactions, which is critical to temporal action proposal task. In our work, we further explore what other specific perturbations are necessary for sequential learning and propose two essential sequential perturbations: temporal feature shift and temporal feature flip. 

The \textbf{temporal feature shift} perturbation is bi-directional moving some randomly selected channels on the feature map of input video along the temporal dimension (Figure~\ref{fig:Network}). Therefore, temporal feature shift can significantly increase the diversity of the input features. Note that, this perturbation is inspired by~\cite{TSM}. The differences between~\cite{TSM} and temporal feature shift include: (1) that~\cite{TSM} chooses fixed channels (select the first $1/4$ of the feature channels, with half moving forward and the other half moving backward). While we randomly choose $\mu$ of feature channels ($\mu$ is a hyper-parameter, $\mu/2$ of feature channels move forward, and the other $\mu/2$ of channels move backward). Hence ours will add more feature diversity. And in the experiments, we observe that the method~\cite{TSM} can lead to a sharp decline in performance since the perturbed training features are completely misaligned compared to the testing features without perturbations. (2) the purpose of ~\cite{TSM} is to achieve the effect of 3D convolution (\emph{i.e.,} to capture the spatio-temporal interactive information between adjacent time points) by inserting this 2D disturbance in residual blocks for action recognition task. 
Our temporal feature shift serves as a way of data augmentation, providing more data for training.

   Besides temporal feature shift, we propose \textbf{temporal feature flip} as another source of sequential perturbation. Since sequential video features with different perturbations may have different numbers of proposals with various locations and sizes, it is challenging to match the given video features. Therefore, the horizontally flipped video features are adopted so that one-to-one correspondence between the proposals in the original and the flipped video features can be easily aligned (Figure~\ref{fig:Network}). During the training, the student models at each iteration are encouraged to generate the symmetric outputs with the teacher models.

   During the training process, each mini-batch includes both labeled and unlabeled data, and we also adopt a dropout strategy to prevent overfitting. The labeled samples are trained using supervised loss. However, without ground truth labels, the supervised loss is undefined upon unlabeled videos. Consistency regularization in mean teacher framework utilizes unlabeled data based on the assumption that the model should output similar predictions when fed perturbed versions of the same input. In our temporal-aware semi-supervised branch, the consistency loss is applied to both the labeled and unlabeled data. Note that, we add consistency loss (L2-loss) to both boundary probability sequences and boundary-matching confidence maps output by BMN. Therefore, in the temporal-aware semi-supervised branch, the total loss formula is:
\begin{equation} \label{eq3} 
{L_{semi}} = L_{supervised} + \lambda_{1}L_{pert\_shift} + \lambda_{2}L_{pert\_flip},
\end{equation}
where weight terms $\lambda_{1}$ and $\lambda_{2}$ are set to 1 and 0.1 separately, $L_{pert\_shift}$ and $L_{pert\_flip}$ are consistency losses for temporal feature shift perturbation and temporal feature flip perturbation separately. 
\subsection{Relation-aware Self-Supervised Branch}
\label{section:3.3}
    Inspired by recent progress in self-supervised learning in video analysis~\cite{video-unsupervised-1,video-unsupervised-2,self-sort-order,self-clip-order,self-PRP}, we hypothesize that the semi-supervised temporal action proposal method could dramatically benefit from self-supervised learning techniques. And based on this insight, in the relation-aware self-supervised branch, we propose two auxiliary tasks. The two auxiliary tasks, \textit{i.e.,} masked feature reconstruction and clip-order prediction, can assist the network in learning temporal relations and discriminative representations.

\noindent \textbf{Masked feature reconstruction. } As shown in Figure~\ref{fig:Network}, the key idea of this self-supervised auxiliary task is to generate the feature $f_{2}$ by randomly masking the video feature $f_{1}$ at some time points along the time dimension. The Base Module then utilizes $f_{2}$ to reconstruct $f_{1}$. The details of the Base Module are shown in Figure~\ref{fig:Base-Module}. Masked feature reconstruction produces self-supervised signals from the original feature $f_{1}$, which can learn discriminative representations in a simple-yet-effective way.  
   
   In the masked feature reconstruction auxiliary task, the Base Module will be driven to perceive and aggregate information from the context to predict the dropped snippets. In this way, the learned temporal semantic relations and discriminative features are conducive to semi-supervised temporal action proposal naturally. We use $\omega$ to represent the degree of the random mask, and we measure the effect of $\omega$ later in Section~\ref{Ablation-Study}.

\noindent \textbf{Clip-order prediction. } This auxiliary task needs to predict clip feature sequences' correct order in a randomly scrambled feature map. Specifically, the reordering of three randomly shuffled feature sequences is shown in Figure~\ref{fig:Network}. Actually, the clip-order prediction is formulated as a classification task. The input is a tuple of clip feature sequences, and the output is a probability distribution over different orders. In the experiment, we empirically designed a reordering of two randomly shuffled feature sequences. The module used for clip-order prediction is shown in Figure~\ref{fig:Base-Module}. 

    Clip-order prediction can leverage the chronological order of feature $f_{1}$ to learn discriminative temporal representations. And clip-order prediction is at the clip sequence level, which can reduce the uncertainty of orders and is more appropriate to learn video feature representations.
\subsection{Overall Loss}
    The total training loss is composed of the losses from section~\ref{section:3.2} and section~\ref{section:3.3}, as follows:
\begin{equation} \label{eq4}
{L_{total}} = L_{semi} + \lambda_{3}L_{aux\_recons} + \lambda_{4}L_{aux\_order},
\end{equation}
where loss functions $L_{aux\_recons}$ and $L_{aux\_order}$ are designed for masked feature reconstruction and clip-order prediction mentioned above separately. Among them,  $L_{aux\_recons}$ is L2-loss and $L_{aux\_order}$ is typical cross-entropy loss for both labeled and unlabeled data. Eventually, the final loss function $L_{total}$ is composed of the $L_{semi}$ in the temporal-aware semi-supervised branch and the losses in the relation-aware self-supervised branch. 
Hyper-parameters $\lambda_{3}$ and $\lambda_{4}$ are set to 0.0001 and 0.001 separately. To jointly learn the semi-supervised pattern and the self-supervised pattern, a unified multi-task framework is exploited for optimization in an end-to-end manner. 
\section{Experiments}

\subsection{Dataset and Setup}
\noindent \textbf{THUMOS14. } This dataset has 1010 validation videos and 1574 testing videos with 20 classes. There are 200 validation videos and 213 testing videos labeled with temporal annotations for the action proposal or detection task. We train our model on the validation set and evaluate on the test set. To make a fair comparison with the previous works~\cite{BMN,ji-Semi}, we employ the same two-stream features~\cite{TSN}.

\noindent \textbf{ActivityNet v1.3. } This dataset is a large-scale dataset containing 19994 videos with 200 activity classes for action recognition, temporal action proposal generation and detection. The quantity ratio of training, validation, and testing sets satisfy 2:1:1. 
Two-stream features are employed to make a fair comparison with the previous works~\cite{BMN,ji-Semi}.
Meanwhile, in order to show that our method is feature-agnostic, we also adopt I3D features~\cite{I3D} pre-trained on Kinetics~\cite{I3D} and without fine-tuned on ActivityNet v1.3. 

We follow the same pre-processing and post-processing steps as the BMN~\cite{BMN}, including parameters adopted in Soft-NMS~\cite{soft-nms} and network structure parameters for a fair comparison.
%
%
\begin{figure}[t]
\begin{center}
\includegraphics[width=0.75\linewidth]{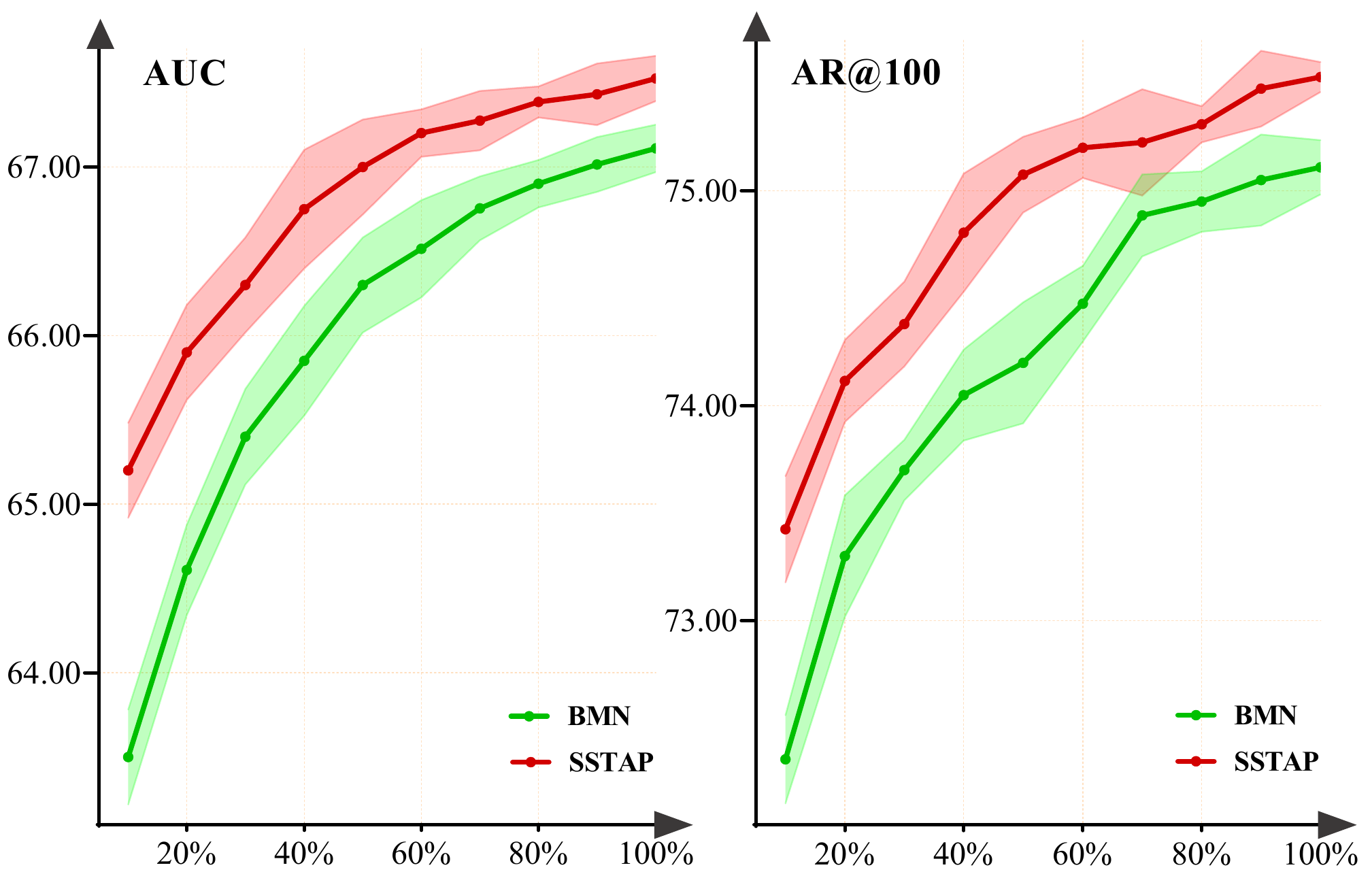}
\vspace*{-5mm}
\end{center}
   \caption{Varying the percentages of labels for training on ActivityNet v1.3, we compare the AUC (left) and AR@100 (right) of the proposals generated by our semi-supervised method and the fully-supervised BMN counterpart.}
\label{fig:Anet-AUC}
\end{figure}
%
%
%
%
%
%
\begin{figure}[t]
\begin{center}
\includegraphics[width=0.95\linewidth]{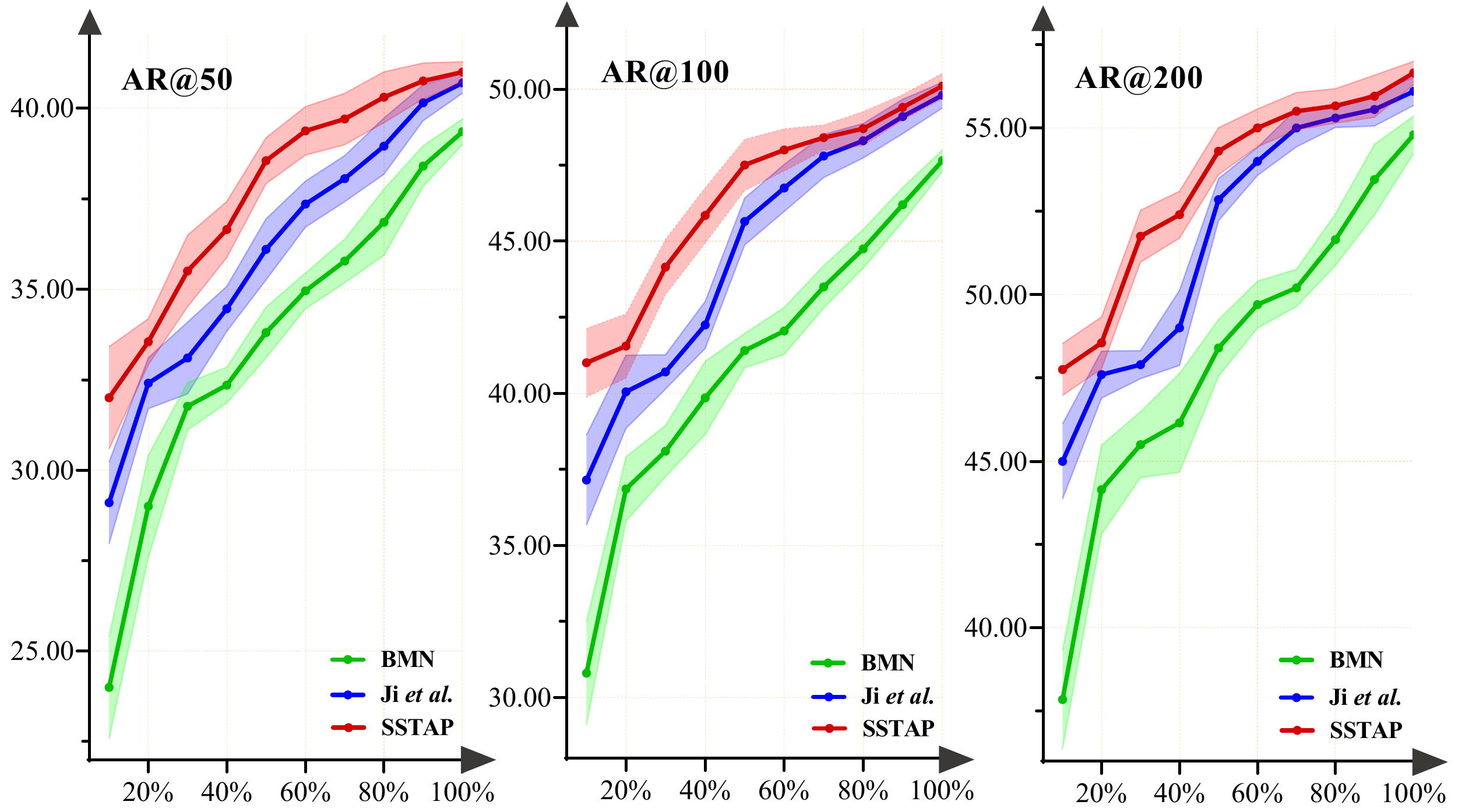}
\vspace*{-5mm}
\end{center}
   \caption{We compare AR@50 (left), AR@100 (middle), and AR@200 (right) of the proposals generated by and the fully-supervised BMN~\cite{BMN}, semi-supervised Ji \etal~\cite{ji-Semi} and our SSTAP when trained with different percentages of labels on the THUMOS14 dataset. Note that, Ji \etal~\cite{ji-Semi} did not publish the source code. For fair comparisons, the results of Ji \etal~\cite{ji-Semi} are carefully reproduced based on BMN~\cite{BMN} by us.}
\label{fig:Thumos14-AR}
\vspace*{-3mm}
\end{figure}
\subsection{Temporal Action Propsal Generation}

   %
   The proposal generation task's goal is to generate high-quality proposals to cover action instances with high recall and high temporal overlap. To evaluate proposal quality, Average Recall (AR) under multiple IoU thresholds are calculated. Following conventions, IoU thresholds $\left [0.5 : 0.05 :0.95\right ]$ and $\left [0.5 : 0.05 : 1.0\right ]$ are used for ActivityNet v1.3 and THUMOS14 respectively. We calculate AR under different Average Number of proposals (AN) as AR@AN and calculate the Area under the AR vs. AN curve (AUC) as metrics on ActivityNet v1.3, where AN is varied from 0 to 100.

\noindent \textbf{Comparsions with fully-supervised methods. } Like~\cite{ji-Semi}, we compare the temporal action proposal results under two training setups: (1) Our semi-supervised framework, where $x\%$ of training videos are labeled with temporal boundaries and $(100-x)\%$ of training videos are unlabeled; (2) Fully-supervised methods, where the same amount of labeled videos are employed for training while no other data are used. With this comparison, we can see how our semi-supervised framework performs against the fully-supervised counterpart under different training ratios.

  We further compare our SSTAP with fully-supervised methods on the validation set of ActivityNet v1.3. Table~\ref{table:Anet-AR-Compare-Supervised} lists a set of proposal generation methods, including TCN~\cite{TCN}, Prop-SSAD~\cite{SSAD}, CTAP~\cite{CTAP}, BSN~\cite{BSN}, MGG~\cite{MGG}, and BMN~\cite{BMN}.
  Specifically, with only 60\% of the videos labeled, our SSTAP surpasses the fully-supervised BMN trained with all labels (100\%) and other fully-supervised methods (Figure~\ref{fig:Anet-AUC} and Table~\ref{table:Anet-AR-Compare-Supervised}). Meanwhile, Table~\ref{table:Anet-AR-Compare-Supervised} also shows that the performance of our SSTAP can be further improved when more labels are available (\emph{i.e.}, 90\% and 100\%). 
  Our approach also performs well with I3D feature inputs, which proves that our SSTAP is feature-agnostic.
  Similarly, Table~\ref{table:Thumos14-AR-Compare-Supervised} and Figure~\ref{fig:Thumos14-AR} show the proposal generation performance comparisons on the testing set of THUMOS14. 
%
%
%
\begin{table}[t]\centering
\subfloat[\# Versus supervised methods.  \label{tab:anet_compare_1}]{
\tablestyle{4pt}{1.0}
\footnotesize
\scalebox{0.83}{
\begin{tabular}{c|cc}
 \small \# Method & \small AR@100 & \small AUC \\
\shline
 TCN & - & 59.58 \\
 Prop-SSAD & 73.01 & 64.40 \\
 CTAP & 73.17 & 65.72\\
 BSN & 74.16 & 66.17 \\
 MGG & 74.54 & 66.43 \\
 SSTAP@60\% & \textbf{75.20} & \textbf{67.23} \\
\end{tabular}}}\hspace{1mm}
\subfloat[\# Versus BMN. \label{tab:anet_compare_2}]{
\tablestyle{3pt}{1.0}
\footnotesize
\scalebox{0.85}{
\begin{tabular}{c|cc}
 \small \# Method & \small AR@100 & \small AUC \\
\shline
 BMN@60\%(I3D) & 74.47 & 66.52 \\
 SSTAP@60\%(I3D) & \textbf{75.00} & \textbf{67.04} \\
 \hline
 BMN@60\% & 74.42 & 66.47 \\
 SSTAP@60\% & \underline{75.20} & \underline{67.23} \\
 BMN@90\% & 74.99 & 67.02 \\
 SSTAP@90\% & \underline{75.46} & \underline{67.48} \\
 BMN@100\% & 75.01 & 67.10 \\
 SSTAP@100\% & \textbf{75.54} & \textbf{67.53} \\
\end{tabular}}}\\\vspace{-3mm}
%
%
\caption{Comparisons between our SSTAP and fully-supervised temporal action proposal generation methods on the validation set of ActivityNet v1.3 dataset in terms of AR@AN and AUC.
\label{table:Anet-AR-Compare-Supervised}}
\vspace{-2mm}
\end{table}
%
%
%
\begin{table}[]
\begin{center}
\scalebox{0.83}{
\begin{tabular}{l|l|cccc}
Feature  & Method     & @50   & @100  & @200  & @1000 \\ 
\shline
2-Stream & TAG~\cite{TAG}        & 18.55 & 29.00 & 39.61 & -     \\
Flow     & TURN~\cite{TURN}       & 21.86 & 31.89 & 43.02 & 64.17 \\
2-Stream & CTAP~\cite{CTAP}       & 32.49 & 42.61 & 51.97 & -     \\
2-Stream & BSN~\cite{BSN}        & 37.46 & 46.06 & 53.21 & 64.52 \\
2-Stream & MGG~\cite{MGG}        & 39.93 & 47.75 & 54.65 & 64.06 \\
2-Stream & DBG~\cite{DBG}        & 37.32 & 46.67 & 54.50 & 66.40 \\
2-Stream & BC-GNN~\cite{BCGNN}     & 40.50 & 49.60 & 56.33 & 66.57 \\
\hline
2-Stream & BMN@60\%   & 34.88 & 42.11 & 49.76 & 61.15 \\
2-Stream & BMN@90\%   & 38.45 & 46.31 & 53.36 & 65.29 \\
2-Stream & BMN@100\%  & 39.36 & 47.72 & 54.70 & 65.49 \\
2-Stream & \textbf{SSTAP@60\%}  & \textbf{39.42} & \textbf{48.02} & \textbf{55.03} & \textbf{67.07} \\
2-Stream & \textbf{SSTAP@90\%}  & \textbf{40.12} & \textbf{49.22} & \textbf{55.86} & \textbf{68.21} \\
2-Stream & \textbf{SSTAP@100\%} & \textbf{41.01} & \textbf{50.12} & \textbf{56.69} & \textbf{68.81} \\ 
\end{tabular}}
\end{center}
\vspace*{-6mm}
\caption{
Comparisons between our method and fully-supervised proposal generation methods on THUMOS14 in terms of AR@AN.
}
\label{table:Thumos14-AR-Compare-Supervised}
\vspace*{-4mm}
\end{table}
\noindent \textbf{{Comparisons with semi-supervised baselines.}} Table~\ref{table:Thumos14-AR-Compare-Semi} compares semi-supervised proposal generation methods on the testing set of the THUMOS14 dataset. To ensure a fair comparison, we adopt the same video feature and post-processing steps. Table~\ref{table:Thumos14-AR-Compare-Semi} shows that our method using two-stream video features outperforms other semi-supervised methods significantly when the proposal number is set within $\left[50,100,200,500,1000\right ]$. Especially, Figure~\ref{fig:Thumos14-AR} demonstrates that our SSTAP outperforms the strong semi-supervised method in Ji \etal~\cite{ji-Semi} consistently under the different ratios of $labeled/(labeled + unlabeled)$ training videos. Unless otherwise stated, the results of Ji \etal~\cite{ji-Semi} are all based on BMN.  
%
%
%
%
%
%
\begin{table}[]
\begin{center}
\scalebox{0.75}{
\begin{tabular}{l|c|ccccc}
Method        & Label         & @50            & @100           & @200           & @500           & @1000          \\ \shline
Vanilla BMN   & 10\%          & 23.71          & 31.11          & 37.98          & 46.35          & 52.25          \\
Mean Teacher~\cite{mean-teacher}  & 10\%          & 27.95          & 36.27          & 43.42          & 51.68          & 57.28          \\
Pseudo-label~\cite{pseudo-label}  & 10\%          & 26.89          & 35.48          & 42.11          & 50.89          & 55.56          \\
Ji \etal~\cite{ji-Semi} & 10\%          & 29.10          & 37.43          & 45.07          & 52.67          & 57.96          \\
\textbf{SSTAP} & \textbf{10\%} & \textbf{32.33} & \textbf{40.92} & \textbf{48.27} & \textbf{54.99} & \textbf{59.38} \\ \hline
Vanilla BMN   & 60\%          & 34.88          & 42.11          & 49.76          & 56.76          & 61.15          \\
Mean Teacher~\cite{mean-teacher}  & 60\%          & 36.77          & 45.23          & 52.26          & 59.50          & 64.04          \\
Pseudo-label~\cite{pseudo-label}  & 60\%          & 36.46          & 45.43          & 53.08          & 59.94          & 63.93          \\
Ji \etal~\cite{ji-Semi}     & 60\%          & 37.42          & 46.71          & 53.96          & 61.01          & 65.10          \\
\textbf{SSTAP} & \textbf{60\%} & \textbf{39.42} & \textbf{48.02} & \textbf{55.03} & \textbf{62.64} & \textbf{67.07} \\ 
\end{tabular}}
\end{center}
\vspace*{-6mm}
\caption{Comparisons between semi-supervised baselines trained with 10\% and 60\% of the labels. For fair comparisons, semi-supervised baselines are all based on BMN. We report AR at various AN on THUMOS14.} 
\label{table:Thumos14-AR-Compare-Semi}
\vspace{-2mm}
\end{table}
%
%
%
%
%
\begin{table}[]
\begin{center}
\scalebox{0.75}{
\begin{tabular}{l|c|ccccc}
Method                            & Label         & @50           & @100          & @200          & @500          & @1000         \\
\shline
Vanilla BMN                       & 10\%          & 23.71          & 31.11          & 37.98          & 46.35          & 52.25          \\
SSTAP - F                       & 10\%          & 32.07          & 40.52          & 47.88          & 54.59          & 58.77          \\
SSTAP - F - R              & 10\%          & 30.82          & 39.24          & 46.85          & 54.31          & 58.71          \\
SSTAP - F - R - C & 10\%          & 30.23          & 38.75          & 46.12          & 53.96          & 58.16          \\
SSTAP - R - C & 10\%          & 30.80          & 38.96          & 46.31          & 54.28          & 58.23          \\
SSTAP - S - R - C & 10\%          & 29.21          & 37.57          & 45.10          & 52.92          & 57.99          \\
\textbf{SSTAP (ALL)}               & \textbf{10\%} & \textbf{32.33} & \textbf{40.92} & \textbf{48.27} & \textbf{54.99} & \textbf{59.38} \\ \hline
Vanilla BMN                       & 60\%          & 34.88          & 42.11          & 49.76          & 56.76          & 61.15          \\
SSTAP - F                       & 60\%          & 39.26          & 48.00          & 54.95          & 62.07          & 66.65          \\
SSTAP - F - R              & 60\%          & 38.52          & 47.24          & 54.69          & 61.89          & 66.72          \\
SSTAP - F - R - C & 60\%          & 38.04          & 46.71          & 54.35          & 62.17          & 66.51          \\
SSTAP - R - C & 60\%          & 38.57          & 46.89          & 54.48          & 62.35          & 66.83          \\
SSTAP - S - R - C & 60\%          & 37.44          & 46.86          & 54.07          & 61.23          & 65.21          \\
\textbf{SSTAP (ALL)}               & \textbf{60\%} & \textbf{39.42} & \textbf{48.02} & \textbf{55.03} & \textbf{62.64} & \textbf{67.07} \\ 
\end{tabular}}
\end{center}
\vspace*{-6mm}
\caption{Ablation study of the effectiveness of components in our SSTAP on THUMOS14. Abbreviations: F for temporal feature flip, R for masked feature reconstruction, C for clip-order prediction, and S for temporal feature shift.}
\label{table:Thumos14-AR-Ablation-Study}
\vspace*{-3mm}
\end{table}

\subsection{Ablation Study}
\label{Ablation-Study}
   In this section, we present ablation studies of several components of our algorithm. We use different values of hyper-parameters that give the best result for each architectural change. The THUMOS14 dataset is employed in all studies performed in this section. 

\noindent \textbf{{Complementarity between components. }} We further conduct detailed ablation studies to evaluate different components of the proposed framework, including temporal feature shift (S), temporal feature flip (F), clip-order prediction (C), and masked feature reconstruction (R). Ablation studies include the following: \\
\textit{Vanilla BMN} : All of the above four components are discarded.\\
\textit{SSTAP - F} : Only temporal feature flip perturbation is discarded. \\
\textit{SSTAP - F - R} : The temporal feature flip perturbation and masked feature reconstruction auxiliary task are discarded.\\
\textit{SSTAP - F - R - C} : The temporal feature flip perturbation and the two self-supervised auxiliary tasks in the relation-aware self-supervised branch are discarded. \\
\textit{SSTAP - R - C} : The two self-supervised auxiliary tasks in the relation-aware self-supervised branch are discarded. \\ 
\textit{SSTAP - S - R - C} : The temporal feature shift perturbation and the two self-supervised auxiliary tasks in the relation-aware self-supervised branch are discarded.

   Table~\ref{table:Thumos14-AR-Ablation-Study} demonstrates that the four components are complementary in terms of improving performance. In particular, when combined with the four components (\emph{i.e.}, \textit{SSTAP (ALL)}), the best performance is achieved. And the results for \textit{SSTAP - F - R - C} and \textit{SSTAP - S - R - C} show that our single perturbation also performs very well.

\noindent \textbf{{Effectiveness of self-supervised branch. }} As illustrated in Table~\ref{table:Thumos14-AR-Ablation-Study-BMN}, we compare the results of applying clip-order prediction (C) and masked feature reconstruction (R) directly to the original BMN. That shows the effectiveness of the two self-supervised auxiliary tasks for performance improvement. Note that, both labeled and unlabeled data are used for training the auxiliary tasks.

\begin{figure}[t]
\begin{center}
\includegraphics[width=0.95\linewidth]{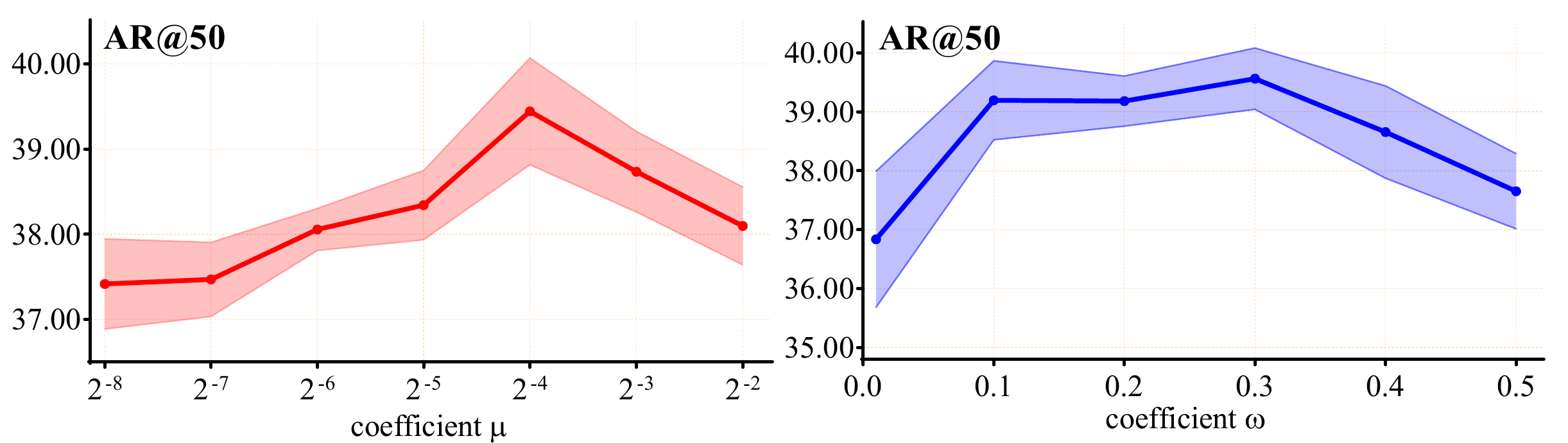}
\vspace*{-5.5mm}
\end{center}
\vspace{-2pt}
   \caption{Ablation comparisons. The effects of temporal feature shift perturbation and masked feature reconstruction auxiliary task under different hyper-parameter choices on the THUMOS14 dataset.}
\label{fig:Ablation_Thumos14}
\vspace*{-3mm}
\end{figure}

\noindent \textbf{{Selection of hyper-parameters. }} Figure~\ref{fig:Ablation_Thumos14} illustrates the comparison of the selection of hyper-parameters. It can be observed that the adjustment of parameters has a certain effect on the performance of AR@50 on THUMOS14, meanwhile, $\mu=2^{-4}$ and $\omega=0.3$ appear to be the optimal operating points.
%
%
%
%
%
\begin{table}[]
\begin{center}
\scalebox{0.79}{
\begin{tabular}{l|c|ccccc}
Method                            & Label         & @50           & @100          & @200          & @500          & @1000         \\
\shline
Vanilla BMN                       & 10\%          & 23.71          & 31.11          & 37.98          & 46.35          & 52.25          \\
BMN + C                       & 10\%          & 26.47          & 34.77          & 41.95          & 49.24       & 54.57          \\
BMN + R                       & 10\%          & 27.45          & 34.89          & 41.51          & 48.71          & 53.75          \\
BMN + C + R              & 10\%          & 28.45          & 36.13          & 42.60          & 49.34          & 54.99          \\
 \hline
Vanilla BMN                       & 60\%          & 34.88          & 42.11          & 49.76          & 56.76          & 61.15          \\
BMN + C                       & 60\%          & 36.75          & 45.76          & 53.05          & 61.78          & 65.84          \\
BMN + R              & 60\%          & 37.14          & 45.83          & 53.17          & 61.20          & 65.75          \\
BMN + C + R & 60\%          & 37.82          & 47.00          & 54.02          & 61.53          & 65.93          \\
\end{tabular}}
\end{center}
\vspace*{-6mm}
\caption{Ablation study of the effectiveness of self-supervised branch. Abbreviations: R for masked feature reconstruction, and C for clip-order prediction.}
\label{table:Thumos14-AR-Ablation-Study-BMN}
\vspace*{-1mm}
\end{table}
%
%
%
%
%
%
%
%
%
\begin{table}[]
\begin{center}
\scalebox{0.75}{
\begin{tabular}{l|c|cccc}
Method    & Reference & 0.5 & 0.75 & 0.95 & Average \\\shline
SCC~\cite{SCC}       & CVPR'17                       & 40.00                   & 17.90                    & 4.70                     & 21.70                       \\
CDC~\cite{CDC}       & CVPR'17                       & 45.30                   & 26.00                    & 0.20                     & 23.80                       \\
R-C3D~\cite{RC3D}     & ICCV'17                       & 26.80                   & -                        & -                        & -                           \\
BSN~\cite{BSN}+~\cite{cuhk}       & ECCV'18                       & 46.45                   & 29.96                    & 8.02                     & 30.03                       \\
TAL-Net~\cite{Rethinking}   & CVPR'18                       & 38.23                   & 18.30                    & 1.30                     & 20.22                       \\
P-GCN~\cite{PGCN}     & ICCV'19                       & 48.26                   & 33.16                    & 3.27                     & 31.11                       \\
G-TAD~\cite{GTAD}+~\cite{cuhk}     & CVPR'20                       & 50.36                   & 34.60                    & 9.02                     & 34.09                       \\
BC-GNN~\cite{BCGNN}+~\cite{cuhk}    & ECCV'20                       & 50.56                   & 34.75                    & \textbf{9.37}                     & 34.26                       \\ \hline
BMN@60\%+~\cite{cuhk}  & ICCV'19                       & 49.50                       & 33.68                        & 8.15 & 33.17                           \\
BMN@90\%+~\cite{cuhk}  & ICCV'19                       & 49.94                       & 33.73                        & 8.23 & 33.74                           \\
BMN@100\%+~\cite{cuhk} & ICCV'19                       & 50.07                   & 34.78                    & 8.29                     & 33.85                       \\
Ji \etal@60\%+~\cite{cuhk}  & ICCV'19                       & 49.82                       & 34.53 & 7.01 & 33.52                           \\
Ji \etal@90\%+~\cite{cuhk}  & ICCV'19                       & 50.24                       & 34.97                        & 7.35 & 34.13                           \\
Ji \etal@100\%+~\cite{cuhk} & ICCV'19                       & 50.55                   & 35.01                    & 7.58                     & 34.23                       \\ \hline
\textbf{SSTAP@60\%}+~\cite{cuhk} & -                             & 50.14                       & 34.92                        & 7.43 & 34.01 \\
\textbf{SSTAP@90\%}+~\cite{cuhk} & -                             & 50.64                       & 35.12                        & 7.80 & 34.35 \\
\textbf{SSTAP@100\%}+~\cite{cuhk} & -                             & \textbf{50.72}                       & \textbf{35.28}                        & 7.87                        & \textbf{34.48}                           \\ 
\end{tabular}}
\end{center}
\vspace*{-5mm}
\caption{Action detection results on the validation set of ActivityNet v1.3, where our proposals are combined with video-level classification results generated by~\cite{cuhk}.}
\label{table:ANet-Det-Compare-SOTA}
\vspace*{-4mm}
\end{table}
\subsection{Action Detection with Our Proposals}
\vspace{-1mm}
   To further examine the quality of the proposals generated by SSTAP, we put the proposals in a temporal action detection framework.
   The evaluation metric of temporal action detection is mAP, which calculates the Average Precision under multiple IoU thresholds for each action category.
   On ActivityNet v1.3, the IoU thresholds for mAP are set to $\{0.5, 0.75, 0.95\}$, and the IoU thresholds for average mAP are set to $\left[0.5 : 0.05 : 0.95\right]$. On THUMOS14, the IoU thresholds for mAP are set to $\{0.3, 0.4, 0.5, 0.6, 0.7\}$.

    We adopt the two-stage ``detection by classifying proposals'' temporal action detection framework to combine our proposals with action classifiers. For fair comparisons, following~\cite{BSN,BMN,GTAD,BCGNN}, on ActivityNet v1.3, we adopt top-1 video-level classification results generated by method~\cite{cuhk} and use confidence scores of BMN proposals for detection results retrieving. On THUMOS14, following BMN~\cite{BMN}, we also use both top-2 video-level classification results generated by UntrimmedNet~\cite{UNet}. And the same classifiers are also used for other proposal generation methods, including SST~\cite{SST}, TURN~\cite{TURN}, BSN~\cite{BSN}, MGG~\cite{MGG}, DBG~\cite{DBG}, G-TAD~\cite{GTAD}, and BC-GNN~\cite{BCGNN}.

   Table~\ref{table:ANet-Det-Compare-SOTA} illustrates the performance comparisons, which are evaluated on the testing set of THUMOS14. With only 60\% of the videos labeled, our SSTAP achieves better performance than fully-supervised BMN trained with all labels in metrics of average mAP. Especially, with 100\% of the videos labeled, our SSTAP outperforms the fully-supervised proposal methods, namely BMN~\cite{BMN}, G-TAD~\cite{GTAD}, BC-GNN~\cite{BCGNN}, and Ji \etal~\cite{ji-Semi}. Similar results on THUMOS14 are shown in Table~\ref{table:Thumos14-AR-Compare-SOTA}, thus demonstrating the effectiveness of our proposed SSTAP.
%
%
\begin{table}[]
\begin{center}
\scalebox{0.75}{
\begin{tabular}{l|c|ccccc}
Method                  & Reference  & 0.7           & 0.6           & 0.5           & 0.4           & 0.3           \\ \shline
SST~\cite{SST}+UNet                & CVPR’17    & 4.7           & 10.9          & 20.0          & 31.5          & 41.2          \\
TURN~\cite{TURN}+UNet               & ICCV’17    & 6.3           & 14.1          & 24.5          & 35.3          & 46.3          \\
BSN~\cite{BSN}+UNet                & ECCV’18    & 20.0          & 28.4          & 36.9          & 45.0          & 53.5          \\
MGG~\cite{MGG}+UNet                & CVPR’19    & 21.3          & 29.5          & 37.4          & 46.8          & 53.9          \\
DBG~\cite{DBG}+UNet                & AAAI’20    & 21.7          & 30.2          & 39.8          & 49.4          & 57.8          \\
G-TAD~\cite{GTAD}+UNet               & CVPR’20    & \textbf{23.4}          & 30.8          & 40.2          & 47.6          & 54.5          \\
BC-GNN~\cite{BCGNN}+UNet             & ECCV’20    & 23.1          & 31.2          & 40.4          & 49.1          & 57.1          \\ \hline
BMN@60\%+UNet           & ICCV’19    & 17.0          & 25.5          & 34.0          & 44.7          & 53.4          \\
BMN@90\%+UNet           & ICCV’19    & 19.7          & 28.9          & 38.2          & 46.8          & 55.5          \\
BMN@100\%+UNet          & ICCV’19    & 20.5          & 29.7          & 38.8          & 47.4          & 56.0          \\ 
Ji \etal@60\%+UNet           & ICCV’19    & 19.2          & 28.1          & 37.1          & 47.2          & 55.4          \\
Ji \etal@90\%+UNet           & ICCV’19    & 21.5          & 31.6          & 41.2         & 50.6          & 57.2          \\
Ji \etal@100\%+UNet          & ICCV’19    & 21.9          & 32.2          & 41.7          & 51.2         & 57.9          \\  \hline
\textbf{SSTAP@60\%+UNet} & - & 20.7 & 30.5 & 39.4 & 48.8 & 56.5 \\
\textbf{SSTAP@90\%+UNet} & - & 22.1 & 32.3 & 41.9 & 51.2 & 57.8 \\
\textbf{SSTAP@100\%+UNet} & - & 22.8 & \textbf{32.8} & \textbf{42.3} & \textbf{51.5} & \textbf{58.4} \\ 
\end{tabular}}
\end{center}
\vspace*{-6mm}
\caption{Action detection results on the testing set of THUMOS14 in terms of mAP@tIoU. We compare with ``proposal + classification'' methods, where classification results are generated by UntrimmedNet~\cite{UNet}.}
\label{table:Thumos14-AR-Compare-SOTA}
\vspace*{-2mm}
\end{table}
\begin{table}[]
\begin{center}
\scalebox{0.81}{
\begin{tabular}{l|c|ccccc}
Method        & Label         & 0.7            & 0.6           & 0.5           & 0.4           & 0.3          \\ \shline
Vanilla G-TAD   & 10\%          & 6.8          & 12.6          & 20.4          & 28.5          & 37.3          \\
Ji \etal~\cite{ji-Semi}+G-TAD     & 10\%          & 9.5          & 17.4          & 25.8          & 34.4          & 43.4          \\
\textbf{SSTAP+G-TAD} & \textbf{10\%} & \textbf{11.1} & \textbf{18.4} & \textbf{27.6} & \textbf{35.9} & \textbf{45.5} \\ \hline
Vanilla G-TAD   & 60\%          & 16.5          & 25.3          & 35.4          & 44.8          & 50.9          \\
Ji \etal~\cite{ji-Semi}+G-TAD    & 60\%          & 20.1          & 29.4          & 39.6          & 47.5          & 53.8          \\
\textbf{SSTAP+G-TAD} & \textbf{60\%} & \textbf{21.8} & \textbf{31.1} & \textbf{41.4} & \textbf{50.2} & \textbf{56.3} \\ \hline
Vanilla G-TAD  & 100\%          & \textbf{23.4}          & 30.8          & 40.2          & 47.6          & 54.5          \\
Ji \etal~\cite{ji-Semi}+G-TAD   & 100\%          & 21.3          & 31.3          & 41.2          & 49.6 & 55.3 \\
\textbf{SSTAP+G-TAD} & \textbf{100\%} & 22.6 & \textbf{32.4} & \textbf{42.7} & \textbf{51.3} & \textbf{57.0} \\ 
\end{tabular}}
\end{center}
\vspace*{-5mm}
\caption{Generalizing our SSTAP to G-TAD~\cite{GTAD} in terms of mAP@tIoU on THUMOS14. The comparison experiments all use the same two-stream feature~\cite{TSN} as in G-TAD~\cite{GTAD}.}
\label{table:Thumos14-GTAD-General}
\vspace*{-3mm}
\end{table}
%
%
%
%
%
%
\subsection{Generalization Experiments}  
  To prove the SSTAP method is valid for other network architectures and frameworks, we introduce SSTAP to G-TAD~\cite{GTAD}. G-TAD proposes to use graph convolutional networks~\cite{GCN} to model temporal relationships between each time point in the input video. As illustrated in Table~\ref{table:Thumos14-GTAD-General}, introducing SSTAP to G-TAD also improves performance. In particular, our SSTAP outperforms the strong semi-supervised baseline~\cite{ji-Semi} by a large margin.
\section{Conclusion} 
In this paper, we incorporate self-supervised learning in the semi-supervised temporal action proposal task and propose a unified SSTAP framework.
%
Specially, we have designed two simple but effective types of temporal sequential perturbations and defined two types of self-supervised pretext tasks for SSTAP.
We show empirically that SSTAP consistently outperforms the state-of-the-art semi-supervised methods and even matches the fully-supervised methods.
Furthermore, we indicate that our SSTAP is agnostic to specific proposal methods and can be effectively applied to other temporal action proposal approaches.
\section*{Acknowledgments}
This work is supported by the National Natural Science Foundation of China under grant (61871435, 61901184),  and the Fundamental Research Funds for the Central Universities no.2019kfyXKJC024.

{\small
\bibliographystyle{ieee_fullname}
\bibliography{egbib}
}

\end{document}